%% file: main.tex
\documentclass{article}

\usepackage{PRIMEarxiv}

\usepackage[utf8]{inputenc} 
\usepackage[T1]{fontenc}    
\usepackage{hyperref}       
\usepackage{url}            
\usepackage{booktabs}       
\usepackage[backend=biber,style=numeric,sorting=none,maxbibnames=5,minbibnames=5]{biblatex}
\usepackage{amsfonts}       
\usepackage{nicefrac}       
\usepackage{microtype}      
\usepackage{lipsum}
\usepackage{fancyhdr}       
\usepackage{graphicx}       
\usepackage{float}
\usepackage{color}
\graphicspath{{media/}{figures/}}     

\newcommand{\roundgenerate}{\textit{generate}}
\newcommand{\roundtune}{\textit{tune}}
\newcommand{\roundevolve}{\textit{evolve}}
\newcommand{\Roundgenerate}{\textit{Generate}}
\newcommand{\Roundtune}{\textit{Tune}}
\newcommand{\Roundevolve}{\textit{Evolve}}

\title{CVEvolve: Zero-code Autonomous Algorithm Discovery for Unstructured Scientific Data Processing}

\author{
  Ming Du$^{*}$, Xiangyu Yin, Yanqi Luo, Dishant Beniwal, Songyuan Tang, Hemant Sharma, Mathew J. Cherukara$^{\dagger}$ \\
  Advanced Photon Source \\
  Argonne National Laboratory \\
  Lemont, IL, USA \\
  $^{*}$\texttt{mingdu@anl.gov}, $^{\dagger}$\texttt{mcherukara@anl.gov} \\
}

\begin{document}

\include{government_license}

\maketitle

\begin{abstract}
Scientific data processing often requires task-specific algorithms or AI models, creating a barrier for domain scientists who need to analyze their data but may not have extensive computing or image-processing expertise. This barrier is especially pronounced when data are noisy, have a high dynamic range, are sparsely labeled, or are only loosely specified. We introduce CVEvolve, an autonomous agentic harness with a zero-code interface for scientific data-processing algorithm discovery. CVEvolve combines a multi-round search strategy with tools for code execution, evaluation implementation, history management, holdout testing, and optional inspection of scientific data and visual outputs. The search alternates between discovery and improvement actions, and uses lineage-aware stochastic candidate sampling to balance exploration and exploitation. We demonstrate CVEvolve on X-ray fluorescence microscopy image registration, Bragg peak detection, high-energy diffraction microscopy image segmentation, and hybrid analytical--learning-based affine registration. Across these tasks, CVEvolve discovers algorithms that improve over baseline methods, while holdout test tracking helps identify candidates that generalize better than later over-optimized alternatives. These results show that zero-code, autonomous LLM-powered algorithm development can help domain scientists turn unstructured scientific image data into practical algorithms and downstream scientific discoveries.
\end{abstract}


\section{Introduction}

Scientific instruments and simulations now produce large volumes of images and other high-dimensional data, and scientific conclusions often depend on the quality of the processing methods applied before interpretation. In practice, this means that researchers routinely need algorithms for tasks such as image registration, denoising, segmentation, peak detection, and feature quantification. Over the last two years, agentic AI systems based on large language models have rapidly expanded from chat-oriented assistants to systems that can plan, use tools, search the literature, write code, and interact with laboratory or computational environments. In science, these capabilities have already been used for autonomous experimentation, hypothesis generation, literature synthesis, and even broader computational research workflows \cite{boiko_autonomous_2023,gottweis_coscientist_2025,skarlinski_synthesis_2024,lu_automation_2026,zheng_autonomy_2025}. A natural next question is whether such systems can go beyond helping scientists implement or use existing data-processing methods and begin to autonomously develop new ones.

AI-assisted research of scientific data-processing algorithms is valuable because this work often falls to domain scientists rather than specialists in computer vision, image processing, or software engineering. In image-heavy fields, even routine analysis can require a long sequence of decisions: which representation to use, how to preprocess data, which assumptions are acceptable, which parameters are stable, and whether a visually plausible output is scientifically meaningful. Prior work in bioimage analysis has repeatedly noted that non-expert users face a knowledge gap in constructing analysis pipelines, selecting tools, and reproducing custom workflows, and that usability and visual feedback are essential for broader adoption of computational methods \cite{ritchie_toolbox_2022,paulgilloteaux_usability_2023}. These observations motivate autonomous systems that do more than write code once: they should be able to inspect data, generate candidates, run them, evaluate them, and revise them in a way that remains understandable to scientists in the loop.

A survey of prior work leads us to group existing systems along three axes: the type of task they target, the type of search harness they use, and the degree to which they support inspection of scientific data and outputs. On the task axis, many influential systems are designed for structured optimization problems. Automated machine learning (AutoML) and neural architecture search methods such as TPOT, neural architecture search, and AutoML-Zero operate in settings with well-defined training and validation data, a constrained design space, and a scalar objective such as classification accuracy or validation loss \cite{olson_tpot_2016,zoph_architecture_2017,real_zero_2020}. More recent LLM-based systems follow the same pattern at a higher level of abstraction. Eureka searches for reward code whose quality is ultimately measured through reinforcement learning performance, and The AI Scientist focuses on machine learning research tasks that still benefit from standard computational benchmarks, code templates, and explicit evaluator loops \cite{ma_eureka_2023,lu_automation_2026}. These works are important steps toward autonomous method development, but they are built around problems whose components, data interfaces, evaluators, and success criteria are relatively well specified.

Scientific data-processing problems are often much less structured. The input may be one image, a folder of image pairs, a diffraction pattern, a set of floating-point arrays, or an acquisition log. The output may be a transformed image, a mask, a list of coordinates, or a derived quantitative measurement. Even when a user can describe the desired performance metric for the output data, turning that description into a reliable executable evaluator can require substantial programming effort. In many cases, visual failure modes also matter: two candidates can obtain similar metrics while producing very different artifacts, false detections, or distortions that are immediately obvious to a human observer but difficult to capture with a single number. This mismatch between the structured evaluator interfaces assumed by many automated method-discovery systems and the messier reality of scientific data processing creates a gap that is especially relevant for domain laboratories.

On the harness axis, existing systems also span a wide range. One family uses relatively direct response--tool--observation loops. Popular coding agents such as OpenAI Codex and Anthropic Claude Code show that language models can iteratively edit files, execute programs, observe outputs, and self-debug in realistic computational environments. These systems are powerful general-purpose coding assistants, but they do not by themselves define a search strategy over families of candidate algorithms. A second family adds an explicit optimization mechanism. FunSearch couples a code-generating model with a programmatic evaluator and an evolutionary population to discover improved programs on formally specified problems \cite{romeraparedes_program_2024}. AlphaEvolve extends this idea to broader scientific and algorithmic problems, while OpenEvolve provides an open-source implementation of this style of evolutionary coding agent \cite{novikov_alphaevolve_2025,sharma_openevolve_2025}. ERA similarly demonstrates that LLM-guided tree search can produce expert-level empirical scientific software when the task is cast as maximizing a quality score \cite{aygun_era_2026}. These systems demonstrate the value of search strategies that explicitly balance exploration and exploitation in sequential decision making. As discussed above, they often assume a scorable-task interface, such as a user-provided executable evaluator, benchmark evaluation code, or prepared validation split. This structure is efficient when the problem interface is already clean, but it gives the agent limited freedom to configure the execution environment, discover underspecified data relationships, repair broken helper scripts, build reusable tools, or preserve intermediate utilities that may become useful in later rounds. Recent systems such as DeepEvolve move beyond this basic loop by incorporating research, richer code modification, debugging, and search memory \cite{liu_deepevolve_2025}. Nevertheless, they still assume a relatively structured problem setup, including an implemented evaluation function and evaluation data, and do not foreground agent-managed workspace preparation as part of the discovery process. For unstructured scientific data-processing tasks, broader workspace-level actions are often part of method development itself rather than peripheral setup. In addition, these methods typically do not provide a mechanism to track the metric on a holdout set, which the development agent is never allowed to see, after each round. For scientific data processing tasks, metric-driven algorithm search can encounter over-optimization (similar to overfitting in data-driven algorithms) to the development data unless additional holdout test mechanisms are introduced.

The third axis is the ability to inspect scientific data and outputs in forms that are meaningful for the task. Direct image viewing is useful for helping the agent understand data characteristics, check intermediate results, and diagnose obvious failure modes. Scientific images can be noisy, contain outliers or anomalous values, exhibit high dynamic ranges, and require preserving the absolute positions of features as well as the quantitative meaning of pixel intensities. They are also often saved in lossless formats such as TIFF, which are rarely accepted directly by LLM APIs. Thus, visualization support is part of the practical infrastructure needed for unstructured scientific data processing. Most existing agentic systems for coding or scientific discovery are fundamentally text-and-code-centric. Recent multimodal coding agents, such as OpenHands-Versa, incorporate visual inputs to expand their ability to interact with diverse environments and improve general task performance, but these systems are not designed as search or optimization frameworks over candidate algorithms \cite{soni_multimodal_2025}. A notable domain-specific exception is PtyChi-Evolve, which uses evolutionary LLM reasoning to discover regularization strategies for ptychography and demonstrates that image-grounded autonomous search can be scientifically useful in a computational imaging setting \cite{yin2026ptychievolve}. This result is encouraging, but it is specialized to one reconstruction problem. A more general framework is still needed for the wider class of scientific data-processing tasks encountered across imaging and beamline science.

Here we introduce CVEvolve, an autonomous framework for scientific data-processing algorithm discovery. CVEvolve is designed for problems where the agent must interact with code, data, metrics, search history, and, when useful, visual outputs in a shared loop, without assuming a predefined problem structure or workflow template. The system operates directly in the space of executable algorithms, including analytical computer vision and image processing methods, and is not restricted to any particular modeling paradigm. It allows users to describe the task, data, and metric in natural language, while the agent writes and runs code, implements evaluation procedures from user-provided metric descriptions, constructs the runtime environment, queries prior search history, inspects generated outputs, and iteratively improves candidate solutions through multiple search actions with different strategic roles. In this way, CVEvolve aims to move autonomous research agents closer to a setting that mirrors how domain scientists actually develop data-processing methods: by combining prior knowledge, computational experiments, quantitative feedback, optional visual inspection, and repeated refinement.

The main contributions of this work are as follows:
\begin{itemize}
\item We present a general agentic framework for autonomous scientific data-processing algorithm discovery that is designed for unstructured problems and does not rely on predefined modeling pipelines or assumptions about data, model, or objective formulations.
\item We provide visualization support that allows the agent to inspect images, plots, and other visual outputs when such inspection is useful during algorithm development. The visualization tools are designed for scientific data, supporting high dynamic range, robustness to outliers and anomalous values, and faithful rendering of quantitative image information.
\item We introduce a long-horizon search harness that combines \roundgenerate{}, \roundtune{}, and \roundevolve{} with lineage-aware state management and an agent-driven holdout test. The lineage-aware design enables structured sampling of parent algorithms based on their evolutionary relationships, promoting both diversity and reuse of successful design patterns during search. The holdout test step is executed by a separate agent that operates on holdout data using only a user-provided data directory and a lightweight description, without requiring predefined holdout test scripts or rigid data formats.
\item We enable the agent to translate user-provided metric descriptions or hints into executable evaluation procedures, removing the need for users to provide a fully specified evaluation harness.
\item We relax the specification of the development and evaluation environment by allowing the agent to construct and manage its own local runtime within the workspace (e.g., via \texttt{uv}), including dependency installation and environment configuration, rather than relying on a pre-configured execution setup.
\item We demonstrate the framework on representative scientific imaging problems and show that it can discover practically useful algorithms that improve substantially over baseline methods.
\end{itemize}

Overall, CVEvolve is a meta-algorithm that, given a task and user instructions, produces candidate algorithms for downstream use within a finite computational budget, while employing an open-ended search process internally. It is intended not as a replacement for domain scientists, but as a tool for lowering the barrier to robust, interpretable, and task-specific data-processing methods.

\section{Results}

\subsection{Algorithm design}

\begin{figure}[H]
  \centering
  \includegraphics[width=1\linewidth]{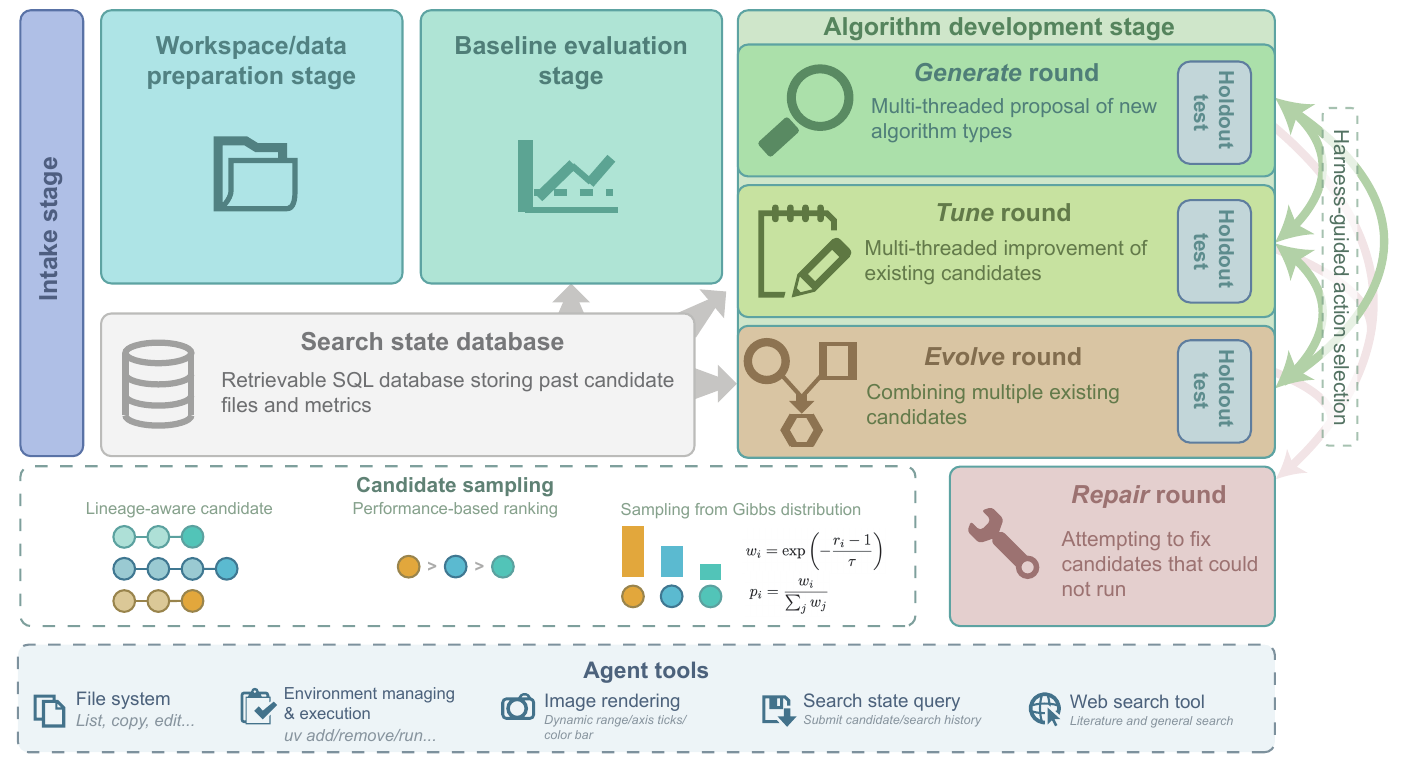}
  \caption{The workflow of CVEvolve. At a high level, the workflow is divided into three stages: in the preparation stage, the agent prepares the workspace, builds the environment, and implements the performance evaluation; in the baseline stage, the agent evaluates user-provided or suggested baseline algorithms; in each round of the algorithm development stage, the harness selects one of the \roundgenerate{}, \roundtune{}, and \roundevolve{} actions with different strategic focuses. For rounds involving parent candidates, the candidates are sampled through a lineage-based approach inspired by MAP-Elites. A holdout test optionally runs at the end of each round on a holdout dataset, using a separate agent to flexibly handle unstructured data and evaluation schemes.}
  \label{fig:workflow}
\end{figure}

\textit{Overall design.} CVEvolve is designed as an autonomous search controller wrapped around an LLM agent that can use code, data, evaluation, history, and visualization tools. In each round, the agent proposes, runs, and evaluates a candidate through tools, and the controller then uses the accumulated history to decide what to try next. Concretely, an algorithm development round chooses from three actions: \roundgenerate{} (proposing new algorithms), \roundtune{} (improving existing algorithms), or \roundevolve{} (hybridizing existing algorithms). This \roundgenerate{}-\roundtune{}-\roundevolve{} strategy is conceptually adapted from the PtyChi-Evolve framework for autonomous ptychography algorithm discovery~\cite{yin2026ptychievolve} and generalized to broader tasks with an expanded toolset and improved state management. To keep the context size under control, the agent starts with a fresh context in each round with only the system prompt and the task prompt specific to the action executed in that round, without inheriting earlier message history. \Roundgenerate{} and \roundtune{} can optionally be executed with multiple parallel workers within the same round, allowing the system to explore several new proposals or several parent-specific refinements concurrently before the history is updated. 

\textit{Candidate and lineage.} At the end of a round, the agent submits the algorithm it developed as a candidate. Candidates are grouped into lineages based on their inheritance relationships. A candidate produced in a \roundgenerate{} round has no parent and becomes the ancestor of its own lineage. A candidate produced in a \roundtune{} round inherits the lineage of its parent. A candidate produced in an \roundevolve{} round starts its own lineage like a \roundgenerate{} candidate, but both of its parents are recorded.

\textit{Search state management.} To support long-horizon search, CVEvolve stores the search history in a persistent relational database that the agent can query through tools during later rounds. The history consists of the candidate pool (including design summaries, analyses, and performance metrics), round history, and metric definitions. This design avoids the drawbacks of traditional memory storage approaches, namely in-context memory and retrieval-augmented generation (RAG) using a vector store. In-context memory does not scale: the context grows across rounds, increases cost, and eventually crowds out information the agent needs for the next decision. It also makes it difficult to reliably access specific fields, such as lineage identifiers or metric values, without repeatedly rewriting large text blocks. RAG with a vector store can help with semantic lookup, but it is not designed for exact, structured record-keeping. Retrieved entries can be incomplete or inconsistent, which weakens reproducibility and complicates deterministic ranking and session recovery.

\textit{Tools.} CVEvolve exposes several tool families to the agent:
\begin{itemize}
    \item \textit{File system tools:} support listing, reading, writing, editing, copying, moving, and deleting files within the workspace, allowing the agent to write candidate code, helper scripts, and evaluation harnesses inside the session sandbox.
    \item \textit{Environment management and code execution tools:} allow candidates to install or remove dependencies in the workspace as well as to execute Python scripts.
    \item \textit{Image viewing tool:} renders images into agent-viewable PNGs. For scientific image data, the rendering tool handles floating-point-valued images and supports percentile-based dynamic-range selection, which suppresses extreme outliers by mapping display bounds to chosen lower and upper intensity percentiles. The tool also supports logarithmic display scaling for high-dynamic-range images. For TIFF images, the tool converts them to PNG images. These controls help the agent inspect weak structures, contrast changes, and failure modes that may be hidden under naive linear rendering.
    \item \textit{Search state tools:} let the agent set the primary metric, log evaluation results, inspect prior candidates and aggregated history, analyze candidate outcomes, record failed attempts, and submit new candidates to the SQL-backed search record.
    \item \textit{Web search tools:} provide access to search APIs for arXiv, Semantic Scholar \cite{kinney2025semanticscholaropendata}, and Tavily so that the agent can retrieve relevant literature and implementation ideas during algorithm development rounds when external technical context is useful.
\end{itemize}
A supporting multimodal detail is the image follow-up middleware: whenever a tool returns an image path in its structured response, the runtime can inject the rendered image back into the conversation as a follow-up message. This mechanism addresses the restriction of common LLM APIs that do not allow tool messages to contain images, enabling the agent to inspect images and plots when visual context is useful.

\textit{Workflow.} The workflow begins with workspace preparation, where the agent inspects copied task data, examines representative images when available, establishes the primary optimization metric from the task description or user hints, and constructs a minimal evaluation harness for later rounds. CVEvolve then runs a baseline round that is reserved for user-provided baselines, directly supplied baseline metrics, or a very small number of explicitly suggested algorithms. Implementing or evaluating user baselines provides benchmarks for later comparison and early candidates for improvement in \roundtune{}/\roundevolve{} rounds. Because the agent resets each round, dedicating a baseline round and storing its results in the search state database prevents redundant reevaluation. After the baseline, CVEvolve enters discovery rounds that follow the \roundgenerate{}-\roundtune{}-\roundevolve{} strategy inherited from PtyChi-Evolve \cite{yin2026ptychievolve} and organized into three categories:
\begin{itemize}
    \item \Roundgenerate{}: broad exploration. The prompt asks the agent to produce a materially new candidate, informed by task characteristics, prior failures, and optionally literature search, while avoiding minor rephrasings of existing code.
    \item \Roundtune{}: exploitative refinement. The agent is given one strong parent candidate sampled from the candidate pool together with its code, metrics, analysis, and settings, and is asked to tune hyperparameters and make fine-grained improvements while keeping the overall design of the algorithm.
    \item \Roundevolve{}: candidate crossover. When multiple candidates exist in the pool, this round samples two parents and prompts the agent to combine their strengths into a new algorithm. If only one candidate is available in the pool, this round falls back to a ``mutate'' action that focuses on improving that single candidate, but the agent is allowed to make more aggressive changes than \roundtune{}.
\end{itemize}
In every round type, the candidate is evaluated in the workspace, analyzed through a structured result-analysis tool, and finally registered through a single formal submission call that stores the candidate artifact path, settings, lineage, and performance summary. 

\textit{Holdout test.} CVEvolve can also perform holdout tests without exposing the holdout data during algorithm design. When this mode is enabled, the main search agent only receives a holdout test prompt that describes the contents of a separate holdout data folder and is instructed to prepare a compact execution contract for later use. After the round ends, CVEvolve materializes a temporary holdout test workspace for the submitted candidate, runs a separate holdout test-only agent on that workspace, records the resulting holdout test metric, and then removes the holdout files from the sandbox.

\textit{Branching mechanism.} The branching mechanism that chooses the next round is history-driven. Early rounds are forced into warmup exploration through \roundgenerate{} actions. After warmup, the controller can periodically force additional \roundgenerate{} rounds to further promote exploration. Otherwise, it first checks whether recent sufficiently strong candidates agree on a suggested next action and, if so, honors that majority signal. If no such suggestion dominates, the controller falls back to scheduled logic based on search progress and performance thresholds: \roundtune{} is selected only when enough excellent candidates exist, while \roundevolve{} is selected only when enough moderate-or-better candidates exist. If neither condition is met, the system falls back to \roundgenerate{}. In this way, action selection combines explicit candidate self-assessment with controller-side safeguards that preserve exploration.

\textit{Stochastic candidate sampling.} For \roundtune{} and \roundevolve{}, CVEvolve uses stochastic candidate sampling rather than always choosing the current best candidate. Our sampling approach is inspired by MAP-Elites \cite{mouret_illuminating_2015}, which is also employed by design search methods like AlphaEvolve \cite{novikov_alphaevolve_2025}. Candidates are first grouped by lineage and ranked by the primary metric. Let $r_i \in \{1,2,\ldots\}$ denote the rank of candidate $i$ in the eligible pool, where $r_i=1$ is best. CVEvolve assigns each candidate an unnormalized selection weight
\begin{equation}
w_i = \exp\left(-\frac{r_i-1}{\tau}\right),
\label{eq:gibbs_sampling}
\end{equation}
where $\tau > 0$ is a temperature parameter controlling exploration. The actual sampling probability is the normalized Gibbs distribution
\begin{equation}
p_i = \frac{w_i}{\sum_j w_j}.
\end{equation}
As $\tau \rightarrow 0^{+}$, the distribution becomes increasingly greedy and concentrates on the top-ranked candidates; larger $\tau$ spreads probability mass more broadly across the ranked pool. When stochastic sampling is disabled, CVEvolve falls back to deterministic top-$k$ parent selection.

For \roundtune{}, the candidate pool for sampling contains one representative per eligible lineage, and parent assignments for multiple workers are sampled without replacement so that distinct workers are encouraged to explore different promising lineages. For crossover in an \roundevolve{} round, CVEvolve constructs a broader pool with multiple candidates per lineage. The first parent is sampled from the distribution above. If one or more parents have already been selected, the weight of a remaining candidate $i$ can be reduced when it belongs to a lineage that has already been chosen:
\begin{equation}
\tilde{w}_i = w_i \, \lambda^{m_i},
\end{equation}
where $\lambda \in [0,1]$ is the same-lineage penalty and $m_i$ is the number of already selected parents sharing candidate $i$'s lineage. The second parent is then sampled from the renormalized distribution based on $\tilde{w}_i$. This mechanism biases crossover toward combining different lineages while still allowing same-lineage crossover when the evidence strongly favors it. Overall, the stochastic lineage-aware sampling broadens the search around multiple promising trajectories instead of collapsing too early onto a single incumbent. Low-level implementation details, including the software framework and concrete runtime stack, are described in the Methods section.

We next present four case studies of CVEvolve on scientific image-processing problems. Claude Opus 4.6 was used for all cases.

\subsection{Case study 1: fluorescence microscopy image registration}

We first demonstrate CVEvolve on the task of finding a robust algorithm for translational registration of X-ray fluorescence microscopy (XRF) images. This problem emerges from the process of optimizing the focus of a fluorescence microscope, where the challenge is that the image drifts spatially once the focusing optics (a Fresnel zone plate) is moved; this necessitates recalibration of the image acquisition position after every move, which requires finding the translation offset between the images before and after the drift. This registration problem is complicated by the size, noise, and varying sharpness of the images. Fig.~S1 in the supplementary document shows an example image pair. These images are simulated by applying shifts, varying Poisson noise, scan jittering, and blurring to a real XRF image, making them representative of the disparities between images that must be registered. The images are plotted on a log scale because the large dynamic range of the image would make most features invisible in a linear-scale plot. The small size of the images (10--30 pixels on each side) restricts the number and resolution of overlapping features that can be anchored on, which particularly affects feature-based registration methods. Meanwhile, the sharpness of the images can vary drastically due to movement of the focusing optics. This inconsistency makes the registration task much more difficult for common strategies such as phase correlation. 

To develop an interpretable and tunable registration algorithm that works without dedicated GPU resources, we tasked CVEvolve with developing an analytical computer vision algorithm for this registration problem. We generated 809 pairs of test/reference images similar to Fig.~S1 and isolated 10\% of them as the holdout set, leaving the rest for algorithm design. 

In the task prompt, we introduced the problem and described the test data directory, which contained the test/reference image pairs and the ground-truth shifts. The test data directory also contained scripts for two baseline algorithms: phase correlation with a Hanning window preprocessor, and a brute-force algorithm that exhaustively searches for the shift that minimizes the test-reference error within a search range. The prompt suggested that the agent use the average Euclidean distance between calculated and ground-truth shifts as the performance metric.

\begin{figure}
  \centering
  \includegraphics[width=0.5\linewidth]{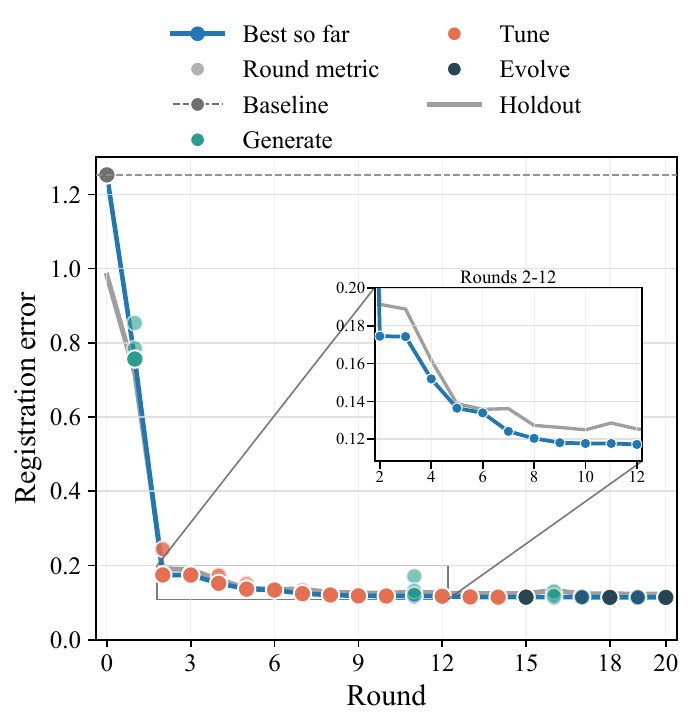}
  \caption{The round types and registration error history throughout the algorithm search for the XRF image registration task. Solid dots represent the best metric up to a round, and dots with lower opacity represent candidates whose metrics did not improve the best record. Inset shows a magnified view of the history between round 2 and 12.}
  \label{fig:xrf_registration_search_history}
\end{figure}

We ran the algorithm search for 20 rounds. The round types and the evolution of the registration error throughout the search are shown in Fig.~\ref{fig:xrf_registration_search_history}. In the baseline evaluation round (round 0), the brute-force method produced an average Euclidean error of 1.25 on the development dataset, while the phase correlation method (not shown) gave an error of 5.8. In the second round, \roundgenerate{} was run, bringing the error down to 0.8, and a subsequent \roundevolve{} round further reduced the error to 0.43. The performance metric started to plateau after round 9. After that, several \roundgenerate{} rounds were run, indicating the agent's attempt to explore the design space and break the plateau. Nevertheless, the error was already low at this stage, and further improvement was bounded by the inherent uncertainty of the data. Fig.~\ref{fig:xrf_registration_search_history} also plots the error on the holdout dataset, which almost coincides with the error on the development dataset. This agreement is expected because the development set is relatively large (though still smaller than the typical size requirement for deep learning methods), and the analytical, parameter-scarce nature of the algorithm makes it less vulnerable to over-optimization. During the run, the agent made several calls to the literature and web search tools. These calls and the use of their results are summarized in Table~S1 in the supplementary document.

To summarize the workflow of the best candidate found, it performs coarse-to-fine image registration by first locating an integer-pixel alignment via multi-scale normalized cross-correlation, then refining to subpixel accuracy using spline/optimizer-based estimation across multiple preprocessing configurations, and finally combining these estimates with adaptive, axis-wise weighting to produce a robust final shift. The candidate was tested on the holdout set and compared with the baseline algorithms in Table~\ref{tab:xrf_registration_holdout}. The error of the best candidate, 0.12, marks a nearly eightfold reduction compared with the better baseline (brute force). This fraction-of-a-pixel average error provides reliable calibration in the XRF auto-focusing process.

We also compared the candidate found by CVEvolve with candidates found by OpenEvolve \cite{sharma_openevolve_2025}, an open-source implementation of
AlphaEvolve \cite{novikov_alphaevolve_2025}, and ERA \cite{aygun_era_2026}. Unlike CVEvolve, OpenEvolve and ERA use a fixed-stage pipeline in each iteration rather than allowing multi-turn, free-form agentic operations within a development round. We therefore ran OpenEvolve and ERA for 500 and 300 iterations, respectively, by which point their metrics had plateaued, and selected the candidate with the best development-set metric from each run. As shown in Table \ref{tab:xrf_registration_holdout}, the OpenEvolve candidate achieved development and holdout errors of 0.20 and 0.23, respectively, compared with 0.11 and 0.12 for the candidate found by CVEvolve. The ERA candidate achieved errors of 0.46 and 0.50. One likely contributor to this difference is CVEvolve's flexible agentic workflow, which allows the agent to inspect task data, construct evaluation code, iterate over candidate designs, and use visualizations as needed within each development round.

\begin{table}
    \centering
    \begin{tabular}{lcc}
    \toprule
    \textbf{Algorithm} & \textbf{Development set} & \textbf{Holdout set} \\
    \midrule
    Brute-force error minimization & 1.25 & 0.98 \\
    Phase correlation & 5.28 & 5.59 \\
    OpenEvolve & 0.20 & 0.23 \\
    ERA & 0.46 & 0.50 \\
    \textbf{CVEvolve} & \textbf{0.11} & \textbf{0.12} \\
    \bottomrule
    \end{tabular}
    \caption{The average euclidean distances of registration results of the baselines and the best candidates found by CVEvolve and prior methods.}
    \label{tab:xrf_registration_holdout}
\end{table}

\subsection{Case study 2: Bragg peak detection}

\begin{figure}[t]
  \centering
  \includegraphics[width=0.95\linewidth]{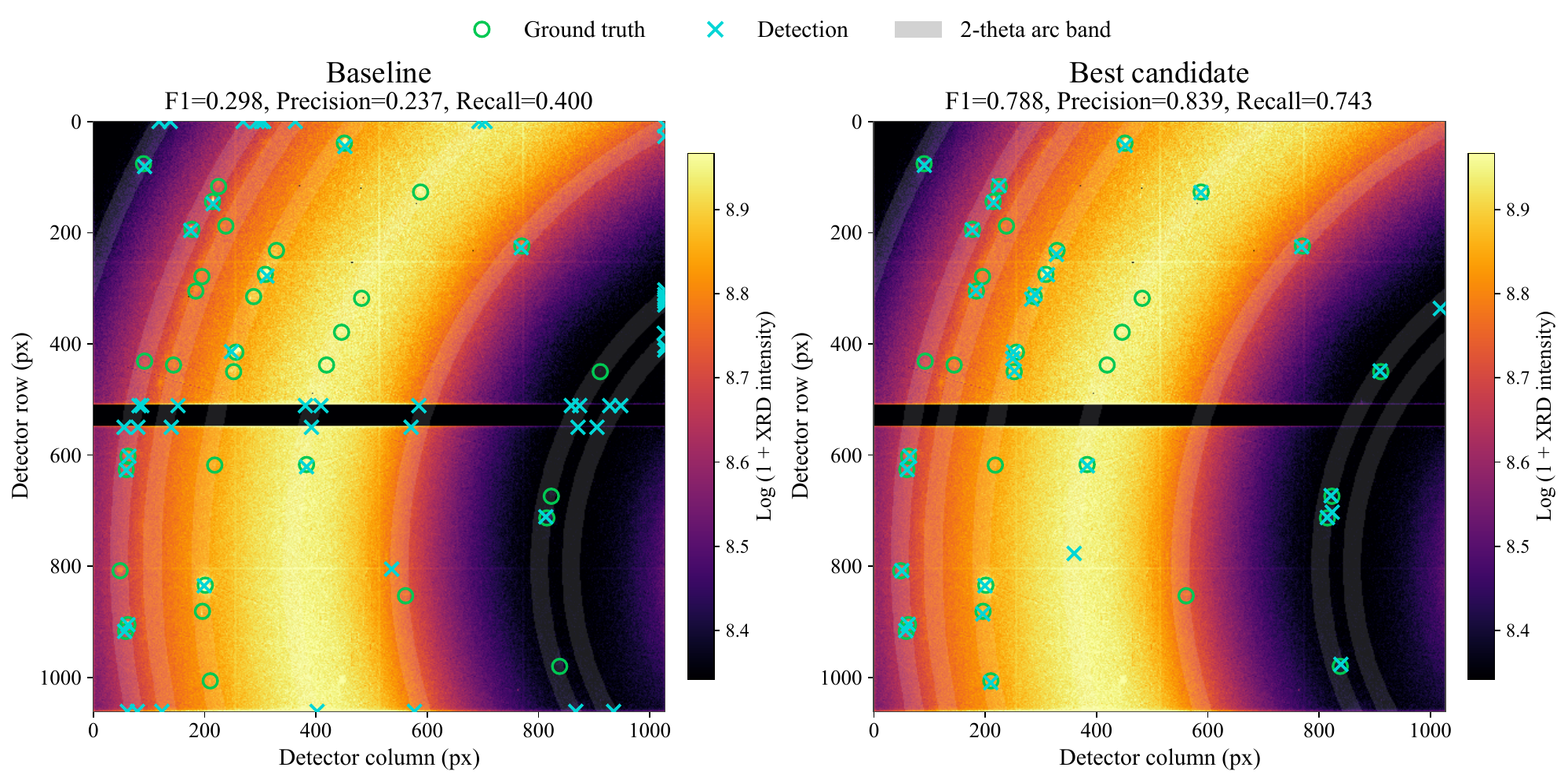}
  \caption{Bragg peak detection results of the baseline and the candidate with the best F1 score on the holdout image. The figure shows the detected peak locations over the diffraction image as well as the manually labeled ground-truth positions.}
  \label{fig:bragg_peak_detection_holdout_result}
\end{figure}

We present another example of using CVEvolve to find an algorithm for Bragg peak detection in X-ray diffraction images. The images are formed by summing the 2D diffraction patterns of a focused X-ray beam across the sample. Bragg peaks are localized bright spots formed by the reflection of X-rays from certain lattice planes of a crystalline sample. For a sample consisting of crystal grains with various orientations, the Bragg peaks associated with a given lattice plane are distributed along a circle in the diffraction image. The objective of this task is to develop a method that identifies and locates Bragg peaks lying on or around the circles corresponding to a series of given lattice planes. The shape, size, and relative intensity of Bragg peaks can vary drastically. The diffraction images also often contain strong and noisy halo-like background, which diminishes the contrast of Bragg peaks in the background regions and adds further challenges to the task (Fig.~\ref{fig:bragg_peak_detection_holdout_result}).

For our experiment, we divided the diffraction images collected from all scan points into two halves and summed the images pixel-wise within each half. This produced two images: one used by the agent to evaluate the algorithm during the development phase, and the other serving as the holdout test image. The Bragg peaks in both images were manually labeled for evaluation.

\begin{figure}[t]
  \centering
  \includegraphics[width=0.6\linewidth]{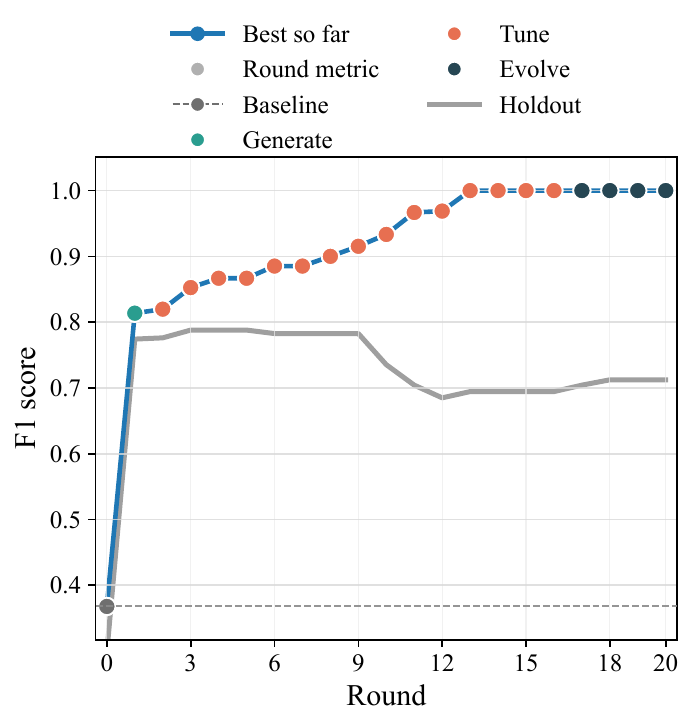}
  \caption{Search history for Bragg peak detection using the hotspot-detection workflow. The figure records the progression of the search process across iterations and provides a visual summary of how candidate hotspot detections evolved during optimization.}
  \label{fig:bragg_peak_detection_search_history}
\end{figure}

Due to the scarcity of the ``development set'' (only one image), CVEvolve can ``overfit'', or over-optimize a candidate to improve the metric on the development dataset. This can occur even though the developed algorithm is an analytical, non-data-driven method, because the agent may make design decisions based on assumptions about sample-specific characteristics or noise structures, or expose too many hyperparameters and tune them to pursue a higher performance metric late in the search. As such, we enabled the holdout test agent in CVEvolve and closely monitored the holdout test metric. In our run, we suggested that the agent use the F1 score as the metric, which balances peak detection and false positive reduction. Fig.~\ref{fig:bragg_peak_detection_search_history} shows the history of the F1 score on both the image used for development-stage evaluation and the one used for holdout testing. The plots reveal that while the F1 score on the development image keeps increasing until it reaches a nearly perfect value of 1, the holdout test F1 score peaks at round 5 and starts to drop sharply after round 9. This observation highlights the importance of holdout testing when the development set is small. Notably, we did not implement any holdout tester specific to the data structure or input/output signatures in this case study; we only provided a directory containing the holdout image, a map of reflection angles for determining the pixels corresponding to a given lattice plane, a list of reflections for peak detection, and a list of true positions, as well as a holdout test prompt describing the contents of the holdout test directory. In the event that the command generated by the development agent is inconsistent with the data, the holdout test agent attempts to fix it so the holdout test runs smoothly without assuming the structure of the problem.

We selected the candidate generated in round 5, which had the best holdout test F1 score. This candidate masks bad regions, performs arc-polar background subtraction and local noise normalization to form an SNR map, then detects peaks through multiple complementary passes (Laplacian of Gaussian/connected components, proximity-based recovery, global faint search, and raw-SNR maxima), followed by merging, prominence/shape validation, and centroid refinement to produce final peak locations. To facilitate understanding of this complex workflow, we present a workflow diagram in Fig.~S2. The peak detection result on the holdout image is shown in Fig.~\ref{fig:bragg_peak_detection_holdout_result}. The left and right panels of the figure show the Bragg peak detection results from the baseline algorithm and the round-5 candidate, respectively. Each plot presents the diffraction image and detected peaks (marked by crosses) along with the manually labeled ground-truth positions. The image contains a blocked region in the middle, whose borders are anomalously bright. Anomalous intensity is also observed at the edges of the image, contributing to many false detections. The best candidate mitigated this problem. Meanwhile, false negatives are also reduced, as more labeled peaks are detected. This improvement is reflected by the higher F1 score (0.298 to 0.788), precision (0.237 to 0.839; associated with false positives), and recall (0.400 to 0.743; associated with false negatives). 

\subsection{Case study 3: diffraction image segmentation}

This section presents an example of a polycrystalline diffraction image segmentation task. Diffraction images of a powder sample may simultaneously exhibit powder rings and Bragg peaks due to reflection from large crystal grains. A significant challenge in this case is that the downstream task requires a strict distinction between powder rings and Bragg peaks. As illustrated in the examples in Fig.~S3, some features appear to be rings, but closer examination reveals that they are in fact formed by overlapping, stretched diffraction peaks with non-uniform width and intensity. These features should be classified as diffraction peaks instead of rings, even though their difference from real powder rings is very subtle. The problem is further complicated by the background of the images, especially the dark ``hole'' at the center due to blockage of X-rays. 

\begin{figure}
    \centering
    \includegraphics[width=0.5\linewidth]{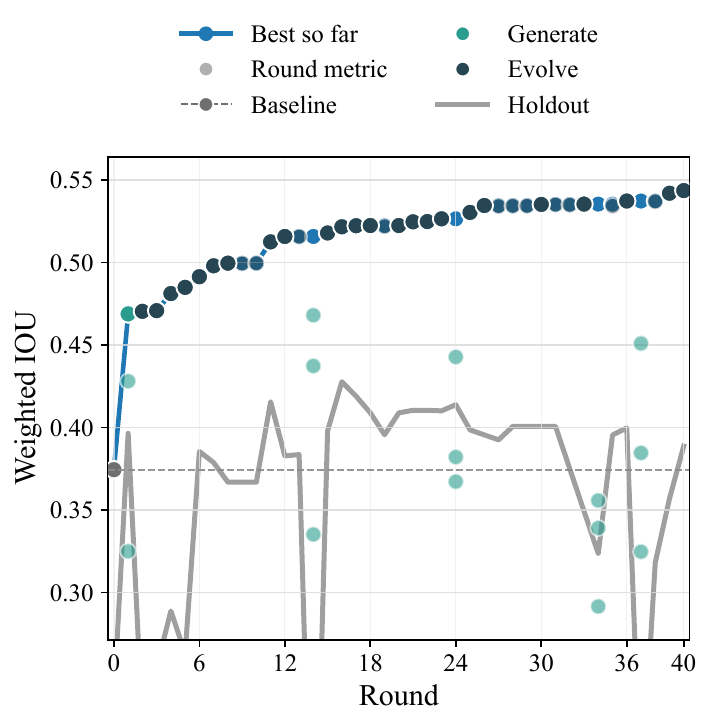}
    \caption{Search history of the HEDM image segmentation task. Solid dots represent the best metric up to a round, and fainter dots represent candidates whose metrics did not improve the best record.}
    \label{fig:hedm_segmentation_search_history}
\end{figure}

For our experiments, we use a weighted intersection-over-union (IoU) score combining the IoU of the segmentation masks for powder rings and diffraction peaks, given as $0.3\mbox{IoU}_\mathrm{ring} + 0.7\mbox{IoU}_\mathrm{peak}$, which weights the accuracy of peaks more heavily due to their relatively smaller footprint. The development dataset contains 5 images and their manually created labels, and the holdout dataset contains 2 samples. We monitored the metric on both the development and holdout datasets over 40 rounds, as shown in Fig.~\ref{fig:hedm_segmentation_search_history}. Although we did not provide any baseline algorithm, the agent at the baseline round (round 0) created a simple baseline candidate that thresholds the image after subtracting a radial background as spots, and labels highly uniform full radial bands as powder rings. This yielded a relatively low IoU of 0.37. 

\begin{figure}
    \centering
    \includegraphics[width=1\linewidth]{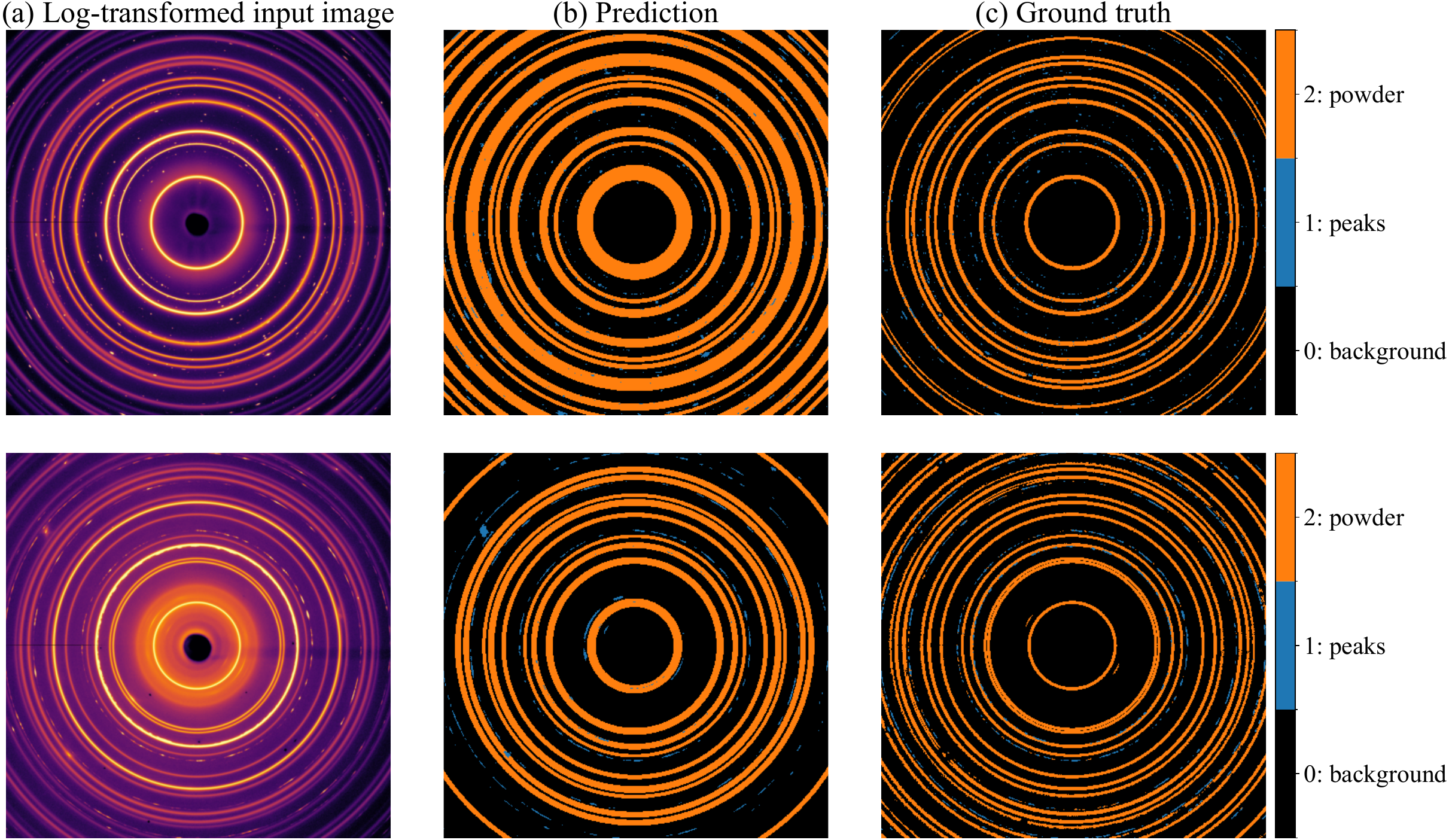}
    \caption{Results of the HEDM segmentation algorithm found by CVEvolve on the holdout dataset. (a) shows the input image transformed with $\log(1 + I)$. (b) and (c) show the predicted and ground-truth segmentation masks, with classes shown in the colorbars.}
    \label{fig:hedm_segmentation_results}
\end{figure}

Due to the high variance across samples, developing a workflow that works for all cases without significant case-specific design or parameter tuning is challenging, and this resulted in a gap between the development and holdout test metrics. However, holdout metric tracking allowed us to identify the candidate with the best holdout metric at round 16. This candidate log-transforms the diffraction image, estimates the beam center and radial background statistics, detects and validates ring structures via radial/azimuthal consistency tests, and then thresholds foreground pixels relative to the background to separate and refine diffraction peaks from rings and background into a final segmentation mask. The result of the candidate on the holdout data is shown in Fig.~\ref{fig:hedm_segmentation_results}. In the first presented case, the predicted ring masks are ``thicker'' than the ground truth, but closer examination confirms that most rings are detected. Individual Bragg peaks are also segmented well, with the predicted mask showing good correlation with the ground truth. In the second case, a few rings in the outer region are missed. However, to present the inherent performance of the candidate found by CVEvolve, we did not perform any per-case parameter tuning, which is often necessary for best results in analytical image processing methods. For Bragg peaks, a few false positives close to the innermost ring are present due to ambiguous signals, and the peaks attached to the inner side of the fourth bright ring from the center are missed. The rest of the Bragg peaks are well segmented, including those with relatively low contrast in the outer region.

\subsection{Case study 4: hybrid analytical and learning-based affine registration}

The case study we present in this section demonstrates a task that is more complicated in implementation: developing a hybrid workflow combining analytical image processing and deep learning (DL) to predict the affine misalignment for cross-resolution image pairs. Accurate and robust registration of cross-resolution image detector systems is a critical step toward several downstream scientific applications. Among them, one example is the fusion of high-resolution and ultra-high-speed camera image sequences \cite{tang_fusion_2025} to investigate highly dynamic processes and phenomena, such as those in the area of additive manufacturing. For the task, the agent was required to set up and run training in its workspace without a human-prepared environment. To add to the challenge, only a small number of images were given to the agent, requiring it to generate more training samples through data augmentation. These complications elevate the task to a complexity similar to a complete data curation and algorithm development cycle of a human-directed DL project.

In the studied problem, an image pair consists of a 1280$\times$800 high-resolution (HR) image and a 400$\times$250 low-resolution (LR) image captured by two different cameras. An exemplar dataset has been uploaded to the TomoBank repository \cite{decarlo_tomobank_2018}, where additional metadata were made available \footnote{\url{https://tomobank.readthedocs.io/en/latest/source/ai/docs.ai.xfusion.html}}. The misalignment between the images is a compound of scale, rotation, and translation. The task is to estimate a two-dimensional scale factor and a subsequent affine transformation that maps the LR image into the coordinate system of the HR image. The development set contains only 15 paired images, while an independent holdout set contains 5 paired images. The task prompt required that the solution include a trained DL model, although it also allowed hybrid methods in which classical registration algorithms are used in sequence with the neural network. Performance was evaluated using target registration error (TRE), defined as the mean Euclidean distance between the predicted and ground-truth transformed positions of the four image corners. The prompt also specified a time budget of 60 minutes for each training run. 

Because the number of available image pairs was small, the agent first divided the development data into 12 seed training pairs and 3 validation pairs, then expanded the effective training set by synthetic augmentation. In particular, training examples were generated by applying random scale, rotation, and translation perturbations to known image pairs, so that the neural network learned to predict residual corrections under a broader range of plausible misalignments. For each seed training pair, 80 augmented pairs were generated, resulting in a total of 960 image pairs as training samples. 

As we did not provide a baseline to the agent, the agent at round 0 developed its own baseline, which first used the scale invariant feature transform (SIFT) algorithm \cite{lowe2004sift} to estimate the transformation and then used a RefineNet model \cite{lin2016refinenet} trained from scratch to predict the residual corrections. This yielded a TRE of 1.74 on the development dataset and 1.54 on the holdout dataset. Additionally, we created a purely analytical workflow that generated SIFT keypoints, matched descriptors with a KD-tree matcher \cite{muja2009flann} followed by ratio-test and RANSAC outlier filtering \cite{fischler1981ransac}, and then applied a second-pass linear residual fitting step to refine scale misalignment. This workflow was applied on the development set separately and resulted in a TRE of 1.81.

\begin{figure}
    \centering
    \includegraphics[width=0.5\linewidth]{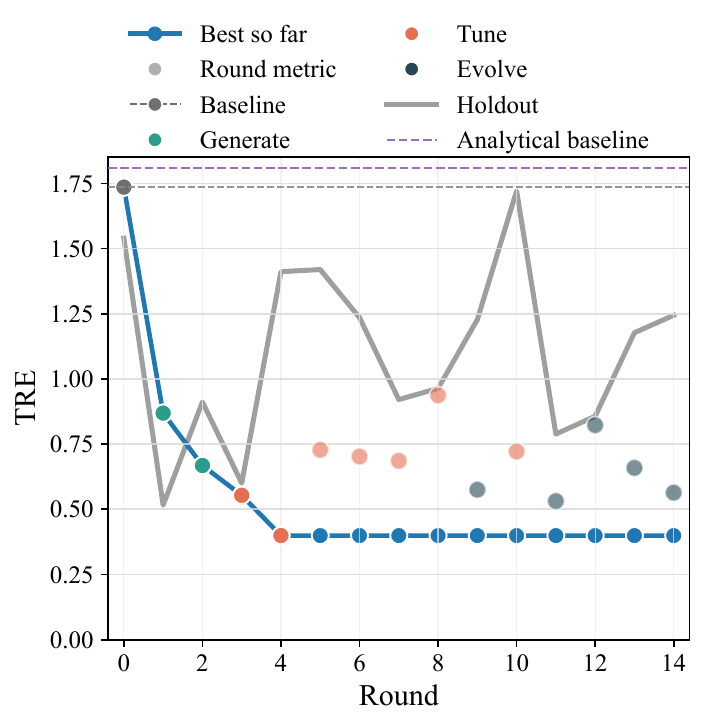}
    \caption{Search history of the hybrid affine registration task. Solid dots represent the best metric up to a round, and fainter dots represent candidates whose metrics did not improve the best record.}
    \label{fig:history_affine_registration}
\end{figure}

The search history is shown in Fig.~\ref{fig:history_affine_registration}. The plot shows both the baseline workflow created by the agent and our crafted analytical workflow. The development TRE reached the best value at round 4, and subsequent rounds were unable to beat that score. The holdout TRE is noisier due to the small data size, but the best scores were also attained before round 4. While the search was set to run for 20 rounds, early stopping occurred at round 14 since no candidates after round 4 yielded a better score. 

Although the holdout TRE at round 1 was the lowest, the metric value is subject to severe statistical noise due to the small size of the holdout set. As such, we chose the candidate at round 3 as the final product, because its development set and holdout set metrics were both low and close to each other, suggesting more stable generalization than candidates whose apparent holdout advantage was not matched by development performance. This candidate used a hybrid registration procedure: SIFT features provided an initial geometric alignment; a Siamese neural network \cite{bromley1993signature} then estimated residual corrections from the aligned image pair; normalized cross-correlation (NCC), a direct measure of image similarity, refined the scale; and Enhanced Correlation Coefficient (ECC) optimization further adjusted the rotation and translation. To improve robustness, the ECC step was run from multiple perturbed initializations and accepted only when it preserved or improved NCC. The holdout TRE of 0.57 is significantly lower than that of the analytical baseline, which is 1.38. We visualize the bounding boxes of the LR images transformed by this candidate and the analytical baseline, and compare them with the ground truth in Fig.~S4. In the presented cases, the transformed boxes of the chosen candidate are always closer to the ground truth compared to the analytical baseline.

\section{Discussion}

CVEvolve uses a stochastic candidate sampling mechanism to introduce exploration into \roundtune{} and \roundevolve{} rounds, preventing them from focusing only on the best-performing candidates seen at that time. To examine the effect of this design, we prepared a ``toy problem'' of finding the maximum of a synthetic numerical function in a 2D space. The function is defined in $0 \le x \le 100$ and $0 \le y \le 100$, and was generated by first randomly sampling 100 points within the space and then constructing a Gaussian kernel density estimator. The values range from 0 to 10, and the maximum is located around (57, 38). CVEvolve was given three initial points: point 1 is closest to the global maximum, but lies in a narrow valley where its value is the lowest among the three; the other two points have higher initial values but are farther away from the global maximum. In a \roundtune{} or \roundevolve{} round, the harness sampled one or two ``parent points'', and the agent was instructed to move from the given point (or from one of the given points) in one of the four cardinal directions. The agent was allowed to make at most four moves in a round, with each move covering a distance no greater than 2. Two workers were used for \roundtune{}. This setup is a simple simulation of a search instance where three baselines are given, one of which has higher potential to evolve into a strong candidate but is not currently performing well; the movement instructions simulate the incremental improvements made to candidates during \roundtune{} or \roundevolve{} rounds.

\begin{figure}
    \centering
    \includegraphics[width=1\linewidth]{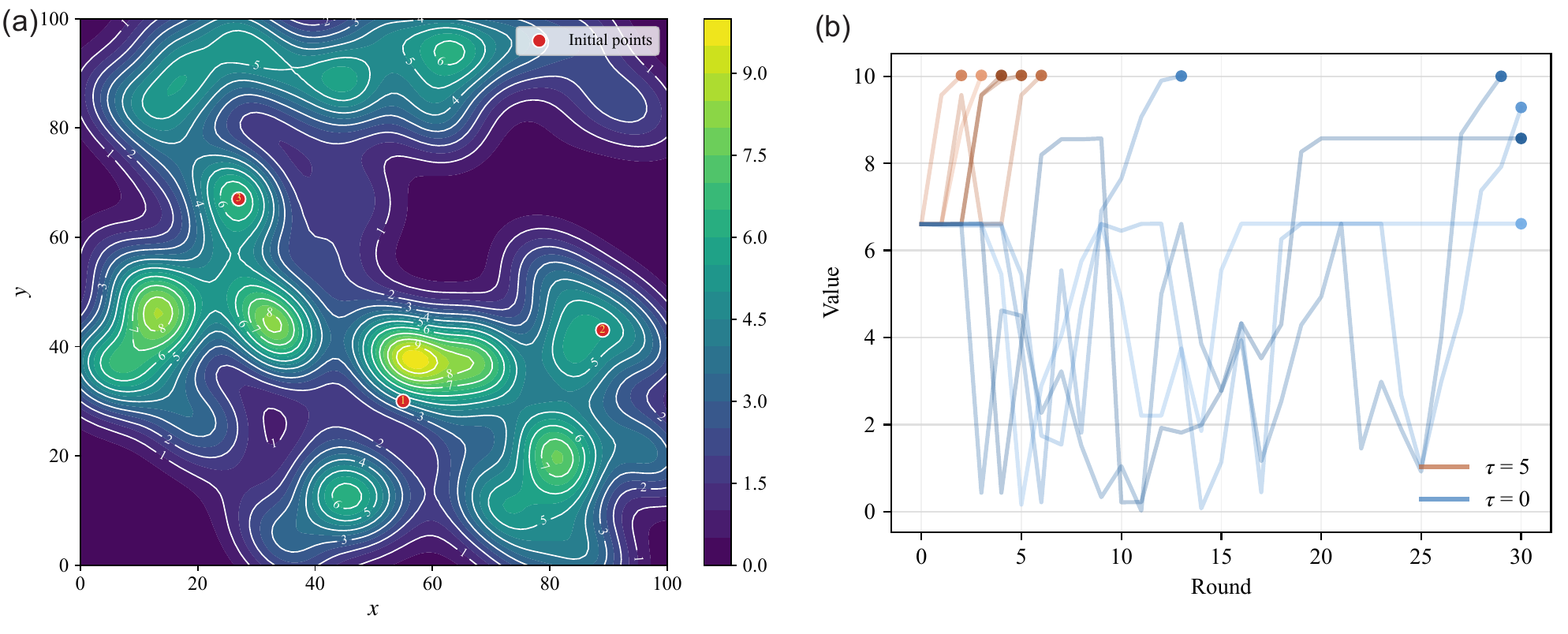}
    \caption{Setup and result of the toy problem. (a) shows the landscape of the search space and the three provided initial points. (b) shows the search history with $\tau = 5$ (red) and $\tau = 0$ (blue), each containing 5 trials.}
    \label{fig:toy_problem}
\end{figure}

We conducted 5 trials with $\tau = 0$ and $\tau = 5$. With a higher temperature $\tau{}$, point 1 and its offspring are more likely to be sampled during a \roundtune{} or \roundevolve{} round despite their lower ``score''. Once selected, the agent can explore in its vicinity, which can lead to fast discovery of the global maximum. If $\tau{}$ is low, candidates that perform well at the starting stage are more likely to be sampled, which results in a higher risk of missing the truly promising candidate (point 1). The result of the experiment shown in Fig.~\ref{fig:toy_problem} is consistent with this analysis: when $\tau = 5$, CVEvolve successfully located the global maximum within 6 rounds in all 5 trials; when $\tau = 0$, only 2 trials found the maximum within the first 30 rounds, and both did so much later, at rounds 13 and 29. Despite being a simple mock test, this experiment mirrors a possible scenario in real algorithm search, where a currently underperforming candidate may in fact be only a few fine-grained tuning steps away from success, and highlights the importance of maintaining a non-zero probability for such candidates to be sampled.

We next discuss the practical security aspects of CVEvolve. To flexibly support unstructured problems, the agents in CVEvolve are given tools to read, copy, move, create, edit, and delete files in the workspace. These tools are guarded by a path checker that detects and rejects operations on files outside the agent's workspace, preventing the agent from unintentionally accessing or modifying files on the host system. However, an important caveat is that because the agent is also given tools to execute Python scripts, it can in fact bypass the guardrail in the file system tools by running file operations in Python. This behavior is difficult for the CVEvolve harness to detect and prevent. The most effective security approach to address such concerns is to deploy CVEvolve in a containerized environment and mount only the workspace directory.

\section{Conclusion}

The rapid increase in the volume and complexity of data generated by modern scientific facilities presents a significant bottleneck for domain researchers. In this paper, we introduced CVEvolve, an autonomous agentic harness for discovering and constructing scientific data-processing algorithms. By avoiding rigid problem templates and allowing users to describe tasks and data in natural language without the need for custom evaluation scripts, while still supporting visual inputs when useful, CVEvolve provides a zero-code interface for algorithm development. This lowers the barrier to entry for computational imaging, enabling domain scientists without extensive computational backgrounds to collaborate with modern AI.

CVEvolve distinguishes itself from prior works that employ fixed, deterministic workflows by granting the agent the freedom to configure the development environment, edit files, and iteratively run tests over multiple turns. It employs a dynamic, multi-round workflow that intelligently selects between \roundgenerate{}, \roundtune{}, and \roundevolve{} strategies based on candidate performance, analysis agent recommendations, and user-specified schedules. Through stochastic candidate sampling inspired by evolutionary search, the harness balances the exploration of novel logic with the exploitation of high-performing code. Furthermore, we address critical vulnerabilities in autonomous algorithm development, including over-optimization and dataset memorization, by introducing an agent-operated holdout test mechanism and harness-enforced data isolation. This provides a reliable way to ensure that the algorithms discovered by CVEvolve generalize robustly.

Empirically, we demonstrated the versatility and performance of CVEvolve across a diverse set of unstructured, real-world scientific imaging tasks, including X-ray fluorescence (XRF) image registration, Bragg peak detection in diffraction patterns, 2D powder diffraction image segmentation, and hybrid affine registration across image resolutions. In these tasks, the agent handled scientific data with floating-point data types, high dynamic ranges, specialized lossless formats (e.g., TIFF), small labeled datasets, and task-specific evaluation needs through a combination of code execution, metric construction, search-state management, holdout testing, optional visualization, and, in the affine-registration case, model training with data augmentation. Ultimately, the harness synthesized algorithms that rival or exceed manual, human-designed baselines.

Looking forward, CVEvolve represents a step toward more autonomous scientific discovery workflows. Future work will focus on expanding the harness to support higher-dimensional data (such as tomography reconstructions and hyperspectral images), \textit{in operando} workflow improvement during live data collections, and the integration of physics-informed constraints into the agent's reasoning process. By shifting part of the burden from manual trial-and-error scripting to autonomous algorithm evolution, systems like CVEvolve can help accelerate the development of practical scientific data-processing methods.

\section{Methods}
\textit{Agent application framework.} CVEvolve is implemented as a LangGraph-based agent application. The runtime uses a compact node graph that separates message ingestion, model reasoning, tool execution, and optional image follow-up handling. When an image path is returned by a tool, the image follow-up node can convert it into a multimodal observation and feed it back to the model before the next reasoning step.

\textit{Environment management and code execution tools.} Candidate evaluation requires controlled access to the Python environment and command execution. To support this, CVEvolve exposes tools that allow the agent to execute commands of the form \texttt{uv} followed by a subcommand and its arguments. In practice, this enables the agent to add dependencies with \texttt{uv add}, remove dependencies with \texttt{uv remove}, and run Python scripts and evaluation harnesses with \texttt{uv run}.

\textit{Holdout test workflow.} CVEvolve supports an optional holdout test through two user inputs: a configured \texttt{holdout\_test\_dir} and a holdout test prompt file passed to the application. The holdout test directory contains data reserved for post-round holdout testing and is not copied into the main search workspace. Instead, the holdout test prompt is injected into the context of algorithm-development agents as a textual description of what the holdout test folder contains. During \roundgenerate{}, \roundtune{}, and \roundevolve{} rounds, these agents therefore know the expected folder structure and file semantics, but they cannot inspect the holdout data themselves. If a runnable candidate is submitted, the agent must also write a \texttt{holdout\_test\_info.json} file in the candidate root. This JSON file specifies three fields: the candidate-root-relative files that must accompany the holdout test, the main execution script among those files, and a \texttt{uv run}-style command intended to execute the candidate in the holdout test workspace with necessary dependencies.

After the round finishes, CVEvolve parses \texttt{holdout\_test\_info.json}, creates a temporary holdout test subdirectory inside the candidate root, copies the holdout test directory into that location, and adds the files listed in \texttt{holdout\_test\_info.json}. A separate holdout test agent is then spawned. The holdout test agent is instructed to use the recorded execution command to run the holdout test, adjusting only environment details or command invocation when needed, while treating the algorithm as read-only. It then submits a numeric holdout test metric or an explicit null result with a failure note. The metric is stored in the database. After the holdout test completes, CVEvolve deletes the temporary holdout test directory immediately so that holdout data do not remain in the sandbox beyond the holdout test step.

\begin{figure}[t]
  \centering
  \includegraphics{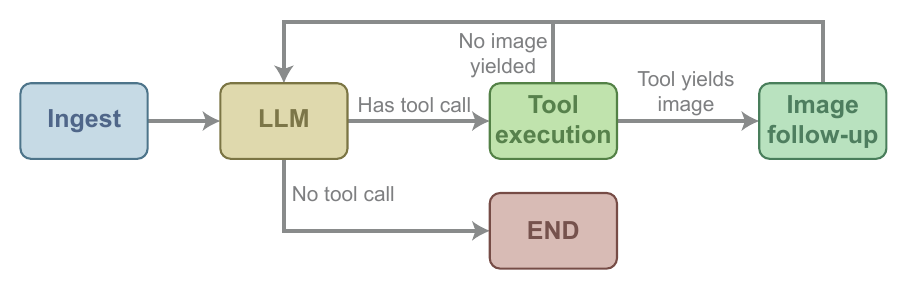}
  \caption{LangGraph execution structure used by CVEvolve. The \textit{Ingest} node prepares the initial prompt (system and round-specific task prompt) and writes it into the state. The state is then passed to the \textit{LLM} node, where the prompt is sent and the LLM response is received. If the response contains a tool call, it is routed to the \textit{Tool execution} node. If the structured return of the tool contains a non-empty image path, it is further routed to the \textit{Image follow-up} node, where the image is loaded, encoded, and appended after the tool response. The tool response and potential image-containing message are routed back to \textit{LLM}. The graph ends if the LLM's response does not contain any tool call.}
  \label{fig:agent_graph}
\end{figure}

\textit{SQL state database.} Search history is stored in a SQLite database that acts as the persistent state backend for the session. The database records rounds, candidate submissions, metric definitions, candidate-level metric values, evaluation metrics, session state, and failure records, and candidate records include lineage information, parent IDs, structured settings, and model-generated performance analysis. In addition to supporting later reporting, this schema serves as the harness' working memory for ranking candidates, estimating performance tiers, tracking lineages, and recovering interrupted sessions. For this reason, candidate submission is treated as part of the search loop rather than passive logging: each successful round formally registers the resulting candidate, its main code file, its parentage, and its structured analysis in the state store.

The database contains the following tables:
\begin{itemize}
    \item \texttt{metric\_definitions}: metric names, optimization directions, descriptions, target values (if given), and the primary-metric flag.
    \item \texttt{rounds}: round index, action type, status, summary, and linked winning candidate.
    \item \texttt{candidates}: submitted candidate records, including round action, description, artifact paths, lineage IDs, parent IDs, notes, and structured metadata.
    \item \texttt{metrics}: candidate-associated metric values used for ranking and comparison.
    \item \texttt{holdout\_test\_metrics}: candidate-associated holdout test metric values.
    \item \texttt{session\_state}: current phase, run status, active round, active action, preparation summary, and stopping reason for session recovery.
    \item \texttt{candidate\_failures}: failed attempts, including failing code path, parent IDs, error message, settings, and metadata.
\end{itemize}

Agent tools related to history retrieval are based on SQL queries with table-joining operations. For example, the \texttt{view\_search\_history}, \texttt{view\_candidate}, and \texttt{view\_metric\_history} tools all rely on cross-table queries. These queries join candidate records with metric tables and, when needed, the primary-metric definition. This allows the tools to recover the latest top-performing candidates, inspect a selected candidate together with its lineage and current score, and map evaluation observations back to candidate identities.

\section*{Acknowledgments}
We thank arXiv for use of its open-access interoperability that supported the literature-search components of this work. We also thank Semantic Scholar and Tavily for API access supporting this search capability. These services were accessed via their publicly available APIs.

This research used resources of the Advanced Photon Source, a U.S.~Department of Energy (DOE) Office of Science user facility at Argonne National Laboratory, and is based on research supported by the U.S. DOE Office of Science-Basic Energy Sciences, under Contract No.~DE-AC02-06CH11357.

This research used Argo, Argonne National Laboratory's internal generative AI interface. Argo was used in CVEvolve to access frontier large language models during our development and experiments.

\section*{Data availability}
The source code of CVEvolve is available at \url{https://github.com/AdvancedPhotonSource/CVEvolve}.

\printbibliography

\end{document}


\maketitle

\section{Example images in the XRF registration problem}

\begin{figure}[H]
    \centering
    \includegraphics[width=0.95\linewidth]{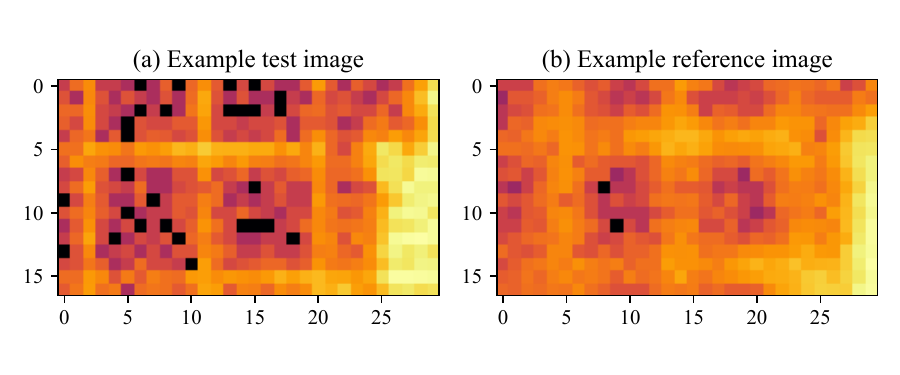}
    \caption{Example test (left) and reference (right) images in the XRF registration problem. The two images are generated by applying different amounts of translational shift, noise, scan jittering, and blurring to a real XRF image in order to simulate images collected under varying focusing and exposure conditions.}
    \label{fig:xrf_registration_example_data}
\end{figure}

\section{Search-tool use in the XRF registration problem}

\begin{table}[H]
  \centering
  \small
  \begin{tabular}{c l p{7.0cm} p{3.5cm}}
  \hline
  \textbf{Round} & \textbf{Search Tool} & \textbf{Query} & \textbf{Result Use} \\
  \hline
  11 & Semantic Scholar
  & \texttt{efficient subpixel image registration DFT upsampling}
  & Survey DFT-based subpixel registration methods \\

  11 & Semantic Scholar
  & \texttt{upsampled DFT cross-correlation sub-pixel image registration}
  & Find Fourier cross-correlation registration references \\

  11 & Semantic Scholar
  & \texttt{efficient subpixel image registration DFT upsampled cross-correlation}
  & Refine research on DFT upsampling methods \\

  11 & Semantic Scholar
  & \texttt{Guizar-Sicairos efficient subpixel registration DFT upsampling 2008}
  & Identify the core Guizar--Sicairos paper \\

  11 & Tavily
  & \texttt{scikit-image phase\_cross\_correlation upsampled DFT sub-pixel registration implementation}
  & Find practical phase-correlation implementation examples \\

  11 & Semantic Scholar
  & \texttt{DOI:10.1364/OL.33.000156}
  & Retrieve metadata for the core paper \\
  \hline
  16 & Semantic Scholar
  & \texttt{optical flow Lucas-Kanade sub-pixel image registration translation estimation}
  & Motivate Lucas--Kanade refinement candidate design \\

  16 & Tavily
  & \texttt{upsampled FFT cross-correlation subpixel image registration Guizar-Sicairos method}
  & Gather implementation references for DFT registration \\

  16 & arXiv
  & \texttt{sub-pixel image registration upsampled cross-correlation matrix-multiply DFT}
  & Search DFT registration literature; results mostly unrelated \\

  16 & Tavily
  & \texttt{scikit-image phase\_cross\_correlation subpixel registration upsampled DFT matrix multiply}
  & Find scikit-image DFT implementation guidance \\
  \hline
  \end{tabular}
  \caption{Literature-search calls in the XRF registration task.}
  \label{tab:xrf_literature_search_calls}
\end{table}

\section{Candidate workflow diagram in the Bragg peak detection problem}

\begin{figure}[H]
    \centering
    \includegraphics[width=0.7\linewidth]{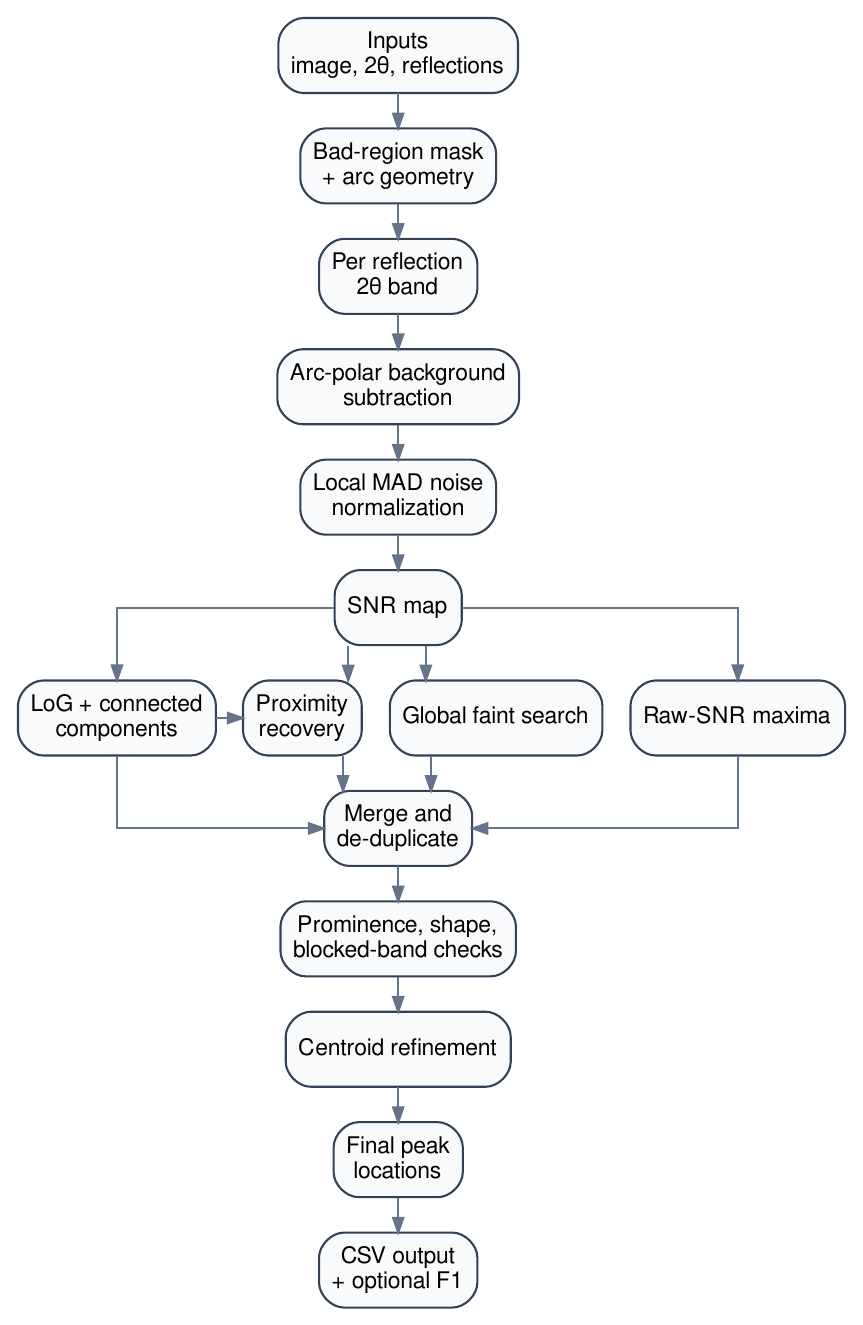}
    \caption{Workflow diagram of the candidate with the best holdout test metric in the Bragg peak detection problem.}
    \label{fig:hotspot_detection_workflow}
\end{figure}

\section{Example images in the HEDM segmentation problem}

\begin{figure}[H]
    \centering
    \includegraphics[width=0.95\linewidth]{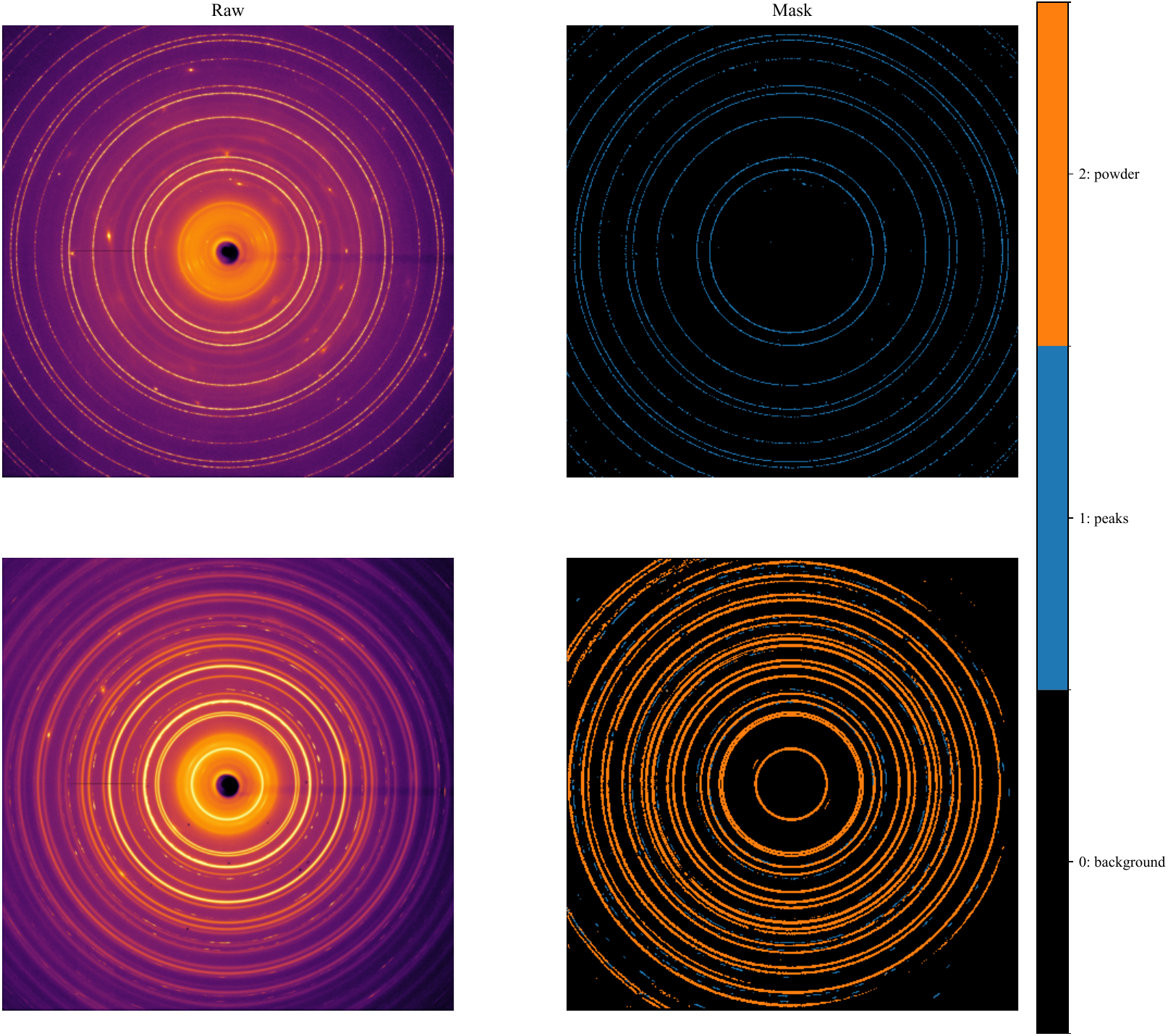}
    \caption{Two examples of the HEDM segmentation problem. The left column shows the raw image, and the right column shows the manually labeled ground truth. There are three segmentation classes: background, diffraction peaks, and powder rings. Some rings, despite appearing circular and continuous at first glance, are in fact composed of overlapping spots, which results in non-uniform width and intensity. The first example illustrates such features, which are labeled as peaks rather than rings.}
    \label{fig:hedm_segmentation_example_data}
\end{figure}

\section{Comparison of baseline and discovered algorithms in the cross-resolution affine image registration problem}

\begin{figure}
    \centering
    \includegraphics[width=1\linewidth]{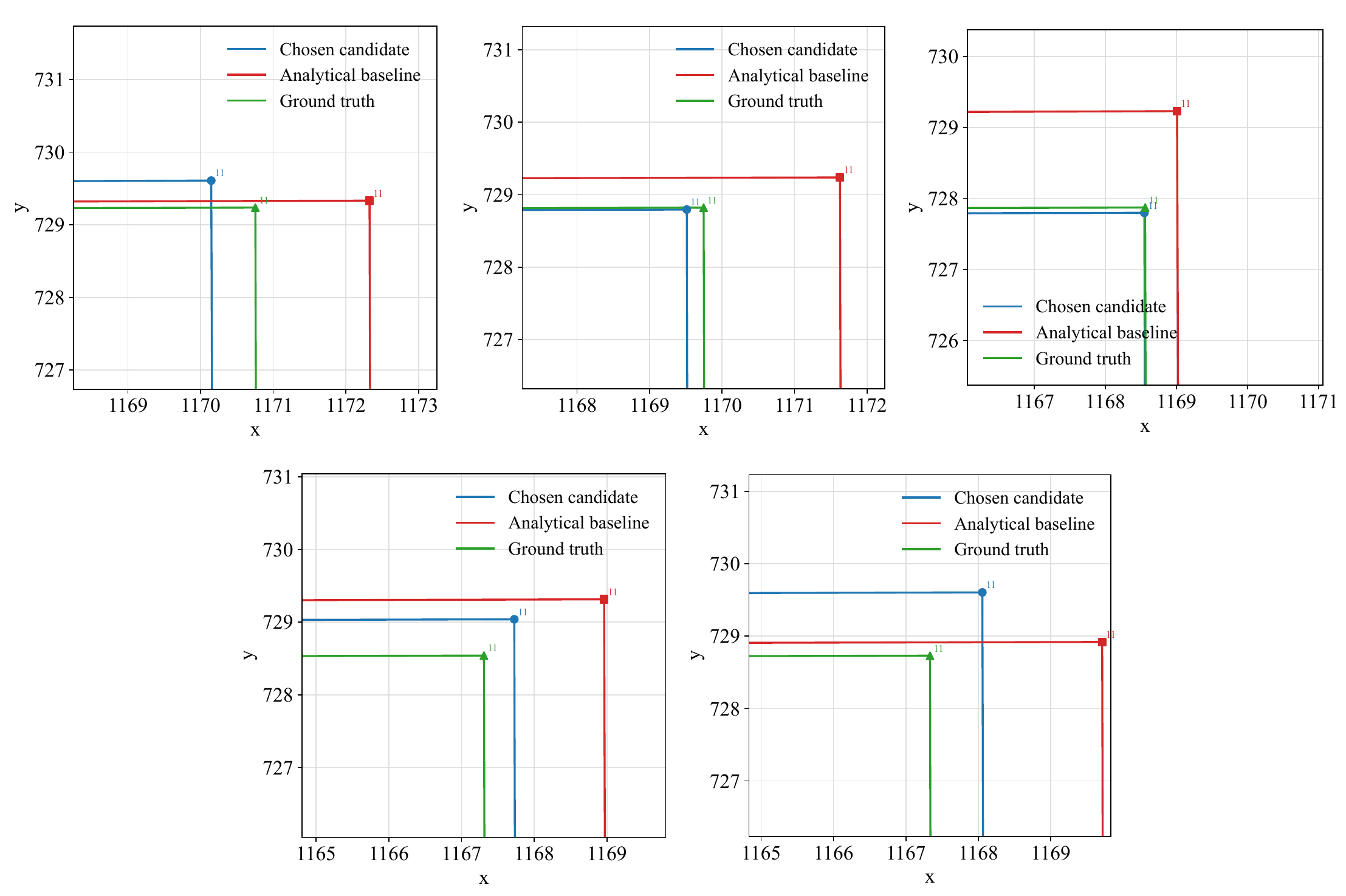}
    \caption{A comparison of the results of the analytical baseline algorithm and the discovered candidate for the cross-resolution affine image registration problem on the holdout dataset. The plots visualize the low-resolution image's bounding box transformed by the ground truth and predicted transformations for each holdout image. To highlight the differences, we zoom in around the transformed top-right corner, originally (400, 250). In the presented cases, the boxes transformed by the chosen candidate (blue) are always closer to the ground truth (green) compared to the analytical baseline (red).}
    \label{fig:affine_registration_holdout_comparison}
\end{figure}


%% file: government_license.tex
\textbf{GOVERNMENT LICENSE}

The submitted manuscript has been created by UChicago Argonne, LLC, Operator of Argonne
National Laboratory (``Argonne''). Argonne, a U.S. Department of Energy Office of Science laboratory, is operated under Contract No. DE-AC02-06CH11357. The U.S. Government retains for
itself, and others acting on its behalf, a paid-up nonexclusive, irrevocable worldwide license in
said article to reproduce, prepare derivative works, distribute copies to the public, and perform
publicly and display publicly, by or on behalf of the Government. The Department of Energy will
provide public access to these results of federally sponsored research in accordance with the DOE
Public Access Plan. http://energy.gov/downloads/doe-public-access-plan.